\def\year{2020}\relax
\documentclass[letterpaper]{article} 
\usepackage{aaai20}  
\usepackage{times}  
\usepackage{helvet} 
\usepackage{courier}  
\usepackage[hyphens]{url}  
\usepackage{graphicx} 
\usepackage{ulem}
\usepackage{booktabs}
\usepackage{amssymb}
\usepackage{amsthm}
\usepackage[tbtags]{amsmath}
\newtheorem{definition}{Definition}
\newcommand*\diff{\mathop{}\!\mathrm{d}}
\DeclareMathOperator*{\argmin}{arg\,min}
\DeclareMathOperator*{\argmax}{arg\,max}
\DeclareMathOperator{\E}{\mathbb{E}}
\usepackage{mathtools}
\DeclarePairedDelimiterX{\infdivx}[2]{(}{)}{%
  #1\;\delimsize\|\;#2%
}
\newcommand{\infdiv}{D_{KL}\infdivx}
\DeclarePairedDelimiter{\norm}{\lVert}{\rVert}
\newcommand{\Tau}{\mathcal{T}}

\usepackage{algorithm}
\usepackage{algpseudocode}
\algnewcommand{\Initialize}[1]{%
  \State \textbf{Initialize:}
  \Statex \hspace*{\algorithmicindent}\parbox[t]{.8\linewidth}{\raggedright #1}
}
\algnewcommand{\Inputs}[1]{%
  \State \textbf{Inputs:}
  \Statex \hspace*{\algorithmicindent}\parbox[t]{.8\linewidth}{\raggedright #1}
}
\algnewcommand{\Outputs}[1]{%
  \State \textbf{Outputs:}
  \Statex \hspace*{\algorithmicindent}\parbox[t]{.8\linewidth}{\raggedright #1}
}
\usepackage{multirow}

\usepackage{color, colortbl}
\definecolor{Gray}{gray}{0.9}
\usepackage[table]{xcolor}

\newcommand{\eat}[1]{}

\usepackage{graphicx}  
\title{Learning from the Past: Continual Meta-Learning with \\Bayesian Graph Neural Networks}
\author{Yadan Luo,\textsuperscript{\rm 1} Zi Huang,\textsuperscript{\rm 1} Zheng Zhang,\textsuperscript{\rm 2,}\textsuperscript{\rm 3}\thanks{Corresponding Author} Ziwei Wang,\textsuperscript{\rm 1} Mahsa Baktashmotlagh,\textsuperscript{\rm 1} Yang Yang\textsuperscript{\rm 4}
\\
\textsuperscript{\rm 1} The University of Queensland, Australia\\ 
\textsuperscript{\rm 2} Bio-Computing Research Center, Harbin Institute of Technology, Shenzhen, China\\
\textsuperscript{\rm 3} Pengcheng Laboratory, Shenzhen, China\\
\textsuperscript{\rm 4} University of Electronic Science and Technology of China, China\\
lyadanluol@gmail.com, huang@itee.uq.edu.au, darrenzz219@gmail.com, \\\{ziwei.wang, m.baktashmotlagh\}@uq.edu.au, dlyyang@gmail.com 
}
\begin{document}

\maketitle

\begin{abstract}
Meta-learning for few-shot learning allows a machine to leverage previously acquired knowledge as a prior, thus improving the performance on novel tasks with only small amounts of data. However, most mainstream models suffer from \textit{catastrophic forgetting} and \textit{insufficient robustness} issues, thereby failing to fully retain or exploit long-term knowledge while being prone to cause severe error accumulation. In this paper, we propose a novel Continual Meta-Learning approach with Bayesian Graph Neural Networks (CML-BGNN) that mathematically formulates meta-learning as continual learning of a sequence of tasks. With each task forming as a graph, the intra- and inter-task correlations can be well preserved via message-passing and history transition. To remedy topological uncertainty from graph initialization, we utilize Bayes by Backprop strategy that approximates the posterior distribution of task-specific parameters with amortized inference networks, which are seamlessly integrated into the end-to-end edge learning. Extensive experiments conducted on the \textit{mini}ImageNet and \textit{tiered}ImageNet datasets demonstrate the effectiveness and efficiency of the proposed method, improving the performance by $42.8\%$ compared with state-of-the-art on the \textit{mini}ImageNet 5-way 1-shot classification task.
\end{abstract}

\section{Introduction}
\noindent 
A key signature of human intelligence is the ability to quickly acquire knowledge from few examples. Despite artificial intelligence has made remarkable progress in wide applications, it remains challenging to perform well in situations with little available data or limited computational resources. Such a scenario is typically referred to as few-shot learning, which has attracted vast interests recently~\cite{aaai1,aaai2,aaai4}. 

Rather than simply augmenting data~\cite{hallucinate} or adding regularization to compensate for the lack of data, an emerging line of work tackles few-shot learning with \textit{meta-learning}. By leveraging previous learning experience to obtain a prior over tasks at meta-train time, the efficiency of later learning is further improved at meta-test time. Particularly, the learned prior for discovering the transferable knowledge can act as an inductive bias to minimize generalization error. 


Generally, mainstream meta-learning models follow the episodic training paradigm, where the \textit{meta-learner} extracts domain-general information among episodes so that it can assist the \textit{task-specific learner} to recognize  unlabeled samples (\textit{query set}) based on the few labeled points (\textit{support set}). In this way, the meta-learner can be implemented variously: as an optimizer that gathers gradient flows from different tasks~\cite{maml,reptile,layer-wise}; as an classification weight generator that hallucinates classifiers for novel classes~\cite{param_prediction,dynamic,gene_GNN}; or as a metric that measures similarity between the query and support samples~\cite{matching,proto}. Nevertheless, existing meta-learning methods are far from optimal due to the lack of \textit{relational inductive bias} modeling~\cite{relational}, thereby failing to manipulate the structured representations of intra- and inter-task relations. 

Motivated to capture more interactions among instances, another line of work has explored graph structure~\cite{GNN,EGNN,TPN} or second-order statistics like covariance~\cite{covariance} in meta-learning framework. More concretely, Garcia and Bruna~\cite{GNN} cast few-shot learning as the node classification problem with graph neural networks, where nodes are represented with the images in the episodes, and edges are given by a trainable similarity kernels. In the same vein, Liu~\cite{TPN} et al. proposed a transductive propagation network (TPN) for label propagation and thus enabled transductive inference for all query set. Alternatively, Kim et al.~\cite{EGNN} modeled the learning as an edge labeling problem, in order to directly predict whether the associated two nodes belong to the same class.

While promising, most existing graph-based meta-learning approaches suffer from two major limitations, \textit{i.e.,} \textit{\textbf{catastrophic forgetting}}~\cite{lifelong} and \textit{\textbf{insufficient robustness}}~\cite{BGNN}, which make it difficult to transfer knowledge over long time spans or handle uncertainty in graph structure. On the one hand, the problem of catastrophic forgetting is that when a meta-learner gradually encounters a sequence of learning problems, it tends to attenuate past knowledge when it learns new things. On the other hand, current graph-based models incorrectly treat the pre-defined edge initialization as the reliable topology for message-passing, in the sense that inaccurate or uncertain relationships among query-support pairs could lead to severe error accumulation through multi-layer propagation.

To address the issues mentioned above, in this paper, we propose a novel Continual Meta-Learning with Bayesian Graph Neural Networks (CML-BGNN) for few-shot classification, which is illustrated in Figure \ref{fig:flowchart}. To alleviate the drawback of catastrophic forgetting, we jointly model the long-term inter-task correlations and short-term intra-class adjacency with the derived continual graph neural networks, which can retain and then access important prior information associated with newly encountered episodes. Specifically, the node update block aggregates the adjacent embedding from each episode and feeds context-aware node representations to gated recurrent units, which are expected to meliorate node features with previous history. Such aggregations can be naturally chained and combined into the multiple layers to enhance model expressiveness. Moreover, as uncertainty is rife in edge initialization, we provide a Bayesian approach for edge inference so that classification weights can be dynamically adjusted for discriminating specific tasks. The conceived amortization networks approximate posterior distribution of classification weights with a Gaussian distribution defined by a mean and variance over possible values. Accordingly, task-specific parameters are sampled to mitigate the bias of node embedding and further enhance the robustness of graph neural networks. Overall, our contributions can be briefly summarized as follows:
\begin{itemize}
    \item We propose a novel Continual Meta-Learning framework that leverages both long-term inter-task and short-term intra-task correlations for few-shot learning. Different from existing graph-based meta-learning approaches, we introduce a memory-augmented graph neural network  to enable flexible knowledge transfer across episodes.
    \item To remedy uncertainty among query-support pairs, a Bayesian edge inference is derived by amortizing posterior inference of task-specific parameters.
    \item We show the effectiveness of the proposed architecture through extensive experiments on the \textit{mini}Imagenet and \textit{tiered}Imagenet benchmark with a $42.8\%$ relative improvement over state-of-the-art counterparts. Regarding to robustness analysis, we perform semi-supervised learning to verify the efficiency and effectiveness of the proposed method.
\end{itemize}

\vspace{-0.5cm}
\section{Related Work}
\subsection{Meta-Learning}
Meta-learning studies how to distill prior knowledge from past experience and enable fast adaptation to novel tasks with only a limited amount of samples. Much effort has been devoted by recent work, which can be broadly categorized into several groups: (1) \textit{Optimization-based} methods either learn a good parameter initialization or leverage an optimizer as the meta-learner to adjust model weights. Typical examples include learning to approximate gradient descent with LSTM~\cite{optimizationasmodel}, learning model-agnostic initial parameters~\cite{maml} and its variants with probabilistic estimation~\cite{pmaml,bmaml,meta-SGD}, first-order approximation~\cite{reptile}, layer selection~\cite{layer-wise}, learner update direction and learning rate learning~\cite{meta-SGD}, and relation embedding~\cite{leo}; (2) \textit{Generation-based} methods learn to augment few-shot data with a generative meta-learner~\cite{PMN}, or learn to predict classification weights for novel classes~\cite{dynamic,param_prediction,gene_GNN}; (3) \textit{Metric-based} approaches address the few-shot classification problem by learning a proper distance metrics as the meta-learner, such as cosine similarity~\cite{matching}, euclidean distance to class prototypes~\cite{proto,softkmeans}, ridge regression~\cite{regression}, relation network~\cite{relation}, task attention~\cite{attention}, category traversal~\cite{ctm}, and graph modeling~\cite{GNN,TPN,covariance,EGNN}. Rather than purely relying on graph-based metric learning, our methodology exploits long-term information from previous tasks and jointly models the topological uncertainty.

\subsection{Catastrophic Forgetting}
Catastrophic forgetting~\cite{lifelong} has been a long-standing issue in machine learning community due to the stability-plasticity dilemma~\cite{dilemma}. In recent literature, a number of methods have been proposed on the basis of continual learning, which can be roughly subdivided into the following groups. Regularization approaches~\cite{regularization} alleviate catastrophic forgetting by imposing constraints on the update of the neural weights in order to prevent ``overwriting" what was previously encoded. Alternatively, in ensemble algorithms~\cite{dynamic}, the architecture itself is altered to accommodate new tasks by retraining a pool of pre-trained models. In dual-memory algorithms~\cite{replay}, one estimates the distribution of the old data either by saving a small fraction of the original dataset into a memory buffer or by training a generator to mimic the lost data and labels~\cite{replay1}. Being more related to the last group, our work, \textit{for the first time}, tackles the catastrophic forgetting in a meta-learning framework and validates its effectiveness and efficiency on practical tasks.

\begin{figure}[t]
    \centering
    \includegraphics[width=1.0\linewidth]{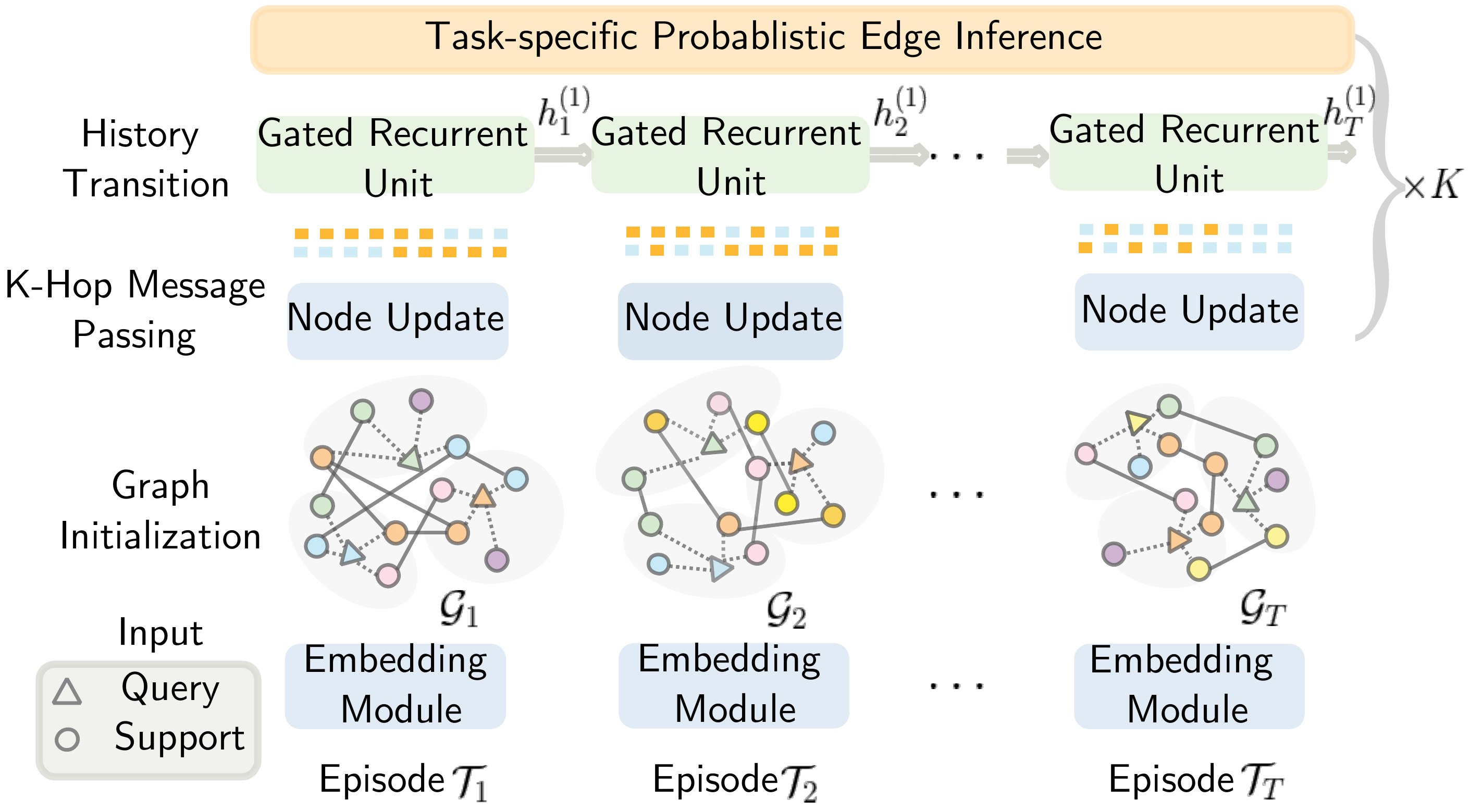}
    \caption{The general flowchart of the proposed continual meta-learning.}
    \label{fig:flowchart}
\end{figure}

\section{Background and Problem Definition}

Given the training set $\mathcal{D}_{train}$, the goal is to learn the model $f : x \rightarrow y$, which is capable of generalizing well to the unseen test set $\mathcal{D}_{test}$, where $\mathcal{D}_{train}\cap\mathcal{D}_{test} = \varnothing$. Meta-learning approaches commonly adopt \textit{episodic training} strategy to minimize the generalization error across a series of tasks $\Tau \sim p(\Tau)$ that are randomly sampled from a task distribution. Specifically, a $N$-way, $K$-shot classification setting is used for both training and testing stage, where $N$ indicates the number of unique classes in each episode and $K$ denotes the number of training samples per class. Each episode or task $\Tau = (\mathcal{S}, \mathcal{Q})$ is composed of the \textit{support set} $\mathcal{S} = \{(x_{i}, y_{i})\}_{i=1}^{N\times K}$ and the \textit{query set} $\mathcal{Q} = \{(\tilde{x}_i, \tilde{y}_i)\}_{i=1}^{K}$. To discover the commonalities and variability across tasks, we decouple the model $f$ into two sub-modules, \textit{i.e.}, the feature learner $f_{\Theta}(\cdot; \Theta)$ and the task-dependent classifiers $\{f_{\psi}(\cdot; \psi_t)\}_{t=1}^{\Tau}$, where $\Theta$ indicates the shared parameters that suit for all tasks and $\{\psi_t\}_{t=1}^{|\Tau|}$ indicate the task-specific parameters. To mathematically illustrate the proposed continual meta-learning procedure, we firstly recall the unified definition of existing meta-learning models: as discussed in literature~\cite{bayes,pmaml},
meta-learning can be viewed as an approximate inference for the posterior given the following definition.
\begin{definition}{\textbf{Meta-Learning}}
Given the task $\Tau$ sampled from task distribution $p(\Tau)$, the posterior predictive distribution for query points is calculated as,

\begin{equation}
    \begin{split}
        p(\widetilde{Y}|\widetilde{X}, \mathcal{S}; \Theta) = &\prod_{i=1}^{|\mathcal{Q}|} p(\tilde{y}_i | \tilde{x}_i, \mathcal{S}_i; \Theta)\\
        = \prod_{i=1}^{|\mathcal{Q}|} \int p (\tilde{y}_i | \tilde{x}_i, &\psi_i; \Theta)p(\psi_i | \mathcal{S}_i; \Theta)\diff\psi_i\approx p(\tilde{y}_i |\tilde{x}_i, \psi_i^*),\nonumber
    \end{split}
\end{equation}
where $\psi^*_i$ is the maximum a posteriori (MAP) value of $\psi_i$, which can be obtained via point estimates.
\end{definition}

 For instance, the \textit{optimization-based} meta-learning regards all model parameters as $\psi$ and forms a point estimate by taking several steps of gradient descent initialized at $\psi_0$ and learning rate $\eta$, \textit{i.e.,}
\begin{equation}
    \psi^*(S; \Theta) = \psi_0 + \eta\frac{\partial}{\partial\psi}\log\sum_{j=1}^{|S|}p(y_j|x_j,\psi;\Theta).
\end{equation}
While, the \textit{generation-based} meta-learning focuses on estimating classification weight vector $\psi$ for novel classes, given the initial value $\psi_0$ trained on $\mathcal{D}_{train}$ and the learning rate $\eta$, \textit{i.e.,}
\begin{equation}
    \psi^*(S,\Theta) = \psi_0 + \eta\frac{\partial}{\partial\psi}\log\sum_{j=1}^{|S|}p(y_j|x_j,G(x_j|\psi); \Theta),
\end{equation}
where $G(\cdot|\cdot)$ is the learnable weight generator.
The \textit{metric-based} meta-learning takes the parameters from the top layer of neural networks as $\psi = \{w_c, b_c\}_{c=1}^C$ for all $C$ classes, by averaging the top-layer activations for each class $c$, \textit{i.e.,}
\begin{equation}
    \begin{split}
    \psi^*(S, \Theta) &= \{\mu_c, -\frac{\parallel \mu_c\parallel^2}{2}\}_{c=1}^{C},\\
    \mu_c &= \frac{1}{k_c}\sum_{j=1}^{k_c}f_{\Theta}(x_j^{(c)}),
    \end{split}
\end{equation}
where $k_c$ denotes the number of samples belonging to class $c$, and $f_{\Theta}(\cdot)$ indicates the embedding network.

However, the current meta-learning suffers from catastrophic forgetting and lacks for uncertainty estimation. Without fixing these problems, a single deep model will be incapable of adapting itself to a long-run learning, since it forgets the old messages when it deals with new things. Therefore, we generalize the meta-learning to a preferable continual way, and give the following elegant definition.
\begin{definition}{\textbf{Continual Meta-Learning}}
Given the task $\Tau$ sampled from task distribution $p(\Tau)$, the posterior predictive distribution for query points is
\begin{equation}
\begin{split}
        p(\widetilde{Y}|\widetilde{X}, \mathcal{S}; \Theta) &= \prod_{i=1}^{|\mathcal{Q}|} p(\tilde{y}_i | \tilde{x}_i, \mathcal{S}_{1:i}; \Theta)\\
        = \prod_{i=1}^{|\mathcal{Q}|} \int p (\tilde{y}_i | \tilde{x}_i, h_i, &\psi_i; \Theta)p(h_i|S_{1:i-1}; \Theta)p(\psi_i | \mathcal{S}_i; \Theta)\diff\psi_i,\nonumber
    \end{split}
\end{equation}
where $h_i$ indicates the history knowledge that gives transition of long-term memory.
\end{definition}

 The first term can be interpreted as given a novel sample $\tilde{x}_i$, its respective label $\tilde{y}_i$ not only conditions on the current task $\psi_i$ but also history information $h_i$, which naturally reuses supervision without pilling up complexity.
The second term presents information updating and resetting from related tasks, which is jointly learned with shared parameter $\Theta$ in the continual graph neural networks, which will be elaborated in the next section. Lastly, to ensure a tractable likelihood, in the last term, the distribution $p(\psi_i|S_i;\Theta)$ over classifier weights is approximated with a few steps of Bayes by Backprop. 

\begin{figure}[t]
    \centering
    \includegraphics[width=1.0\linewidth]{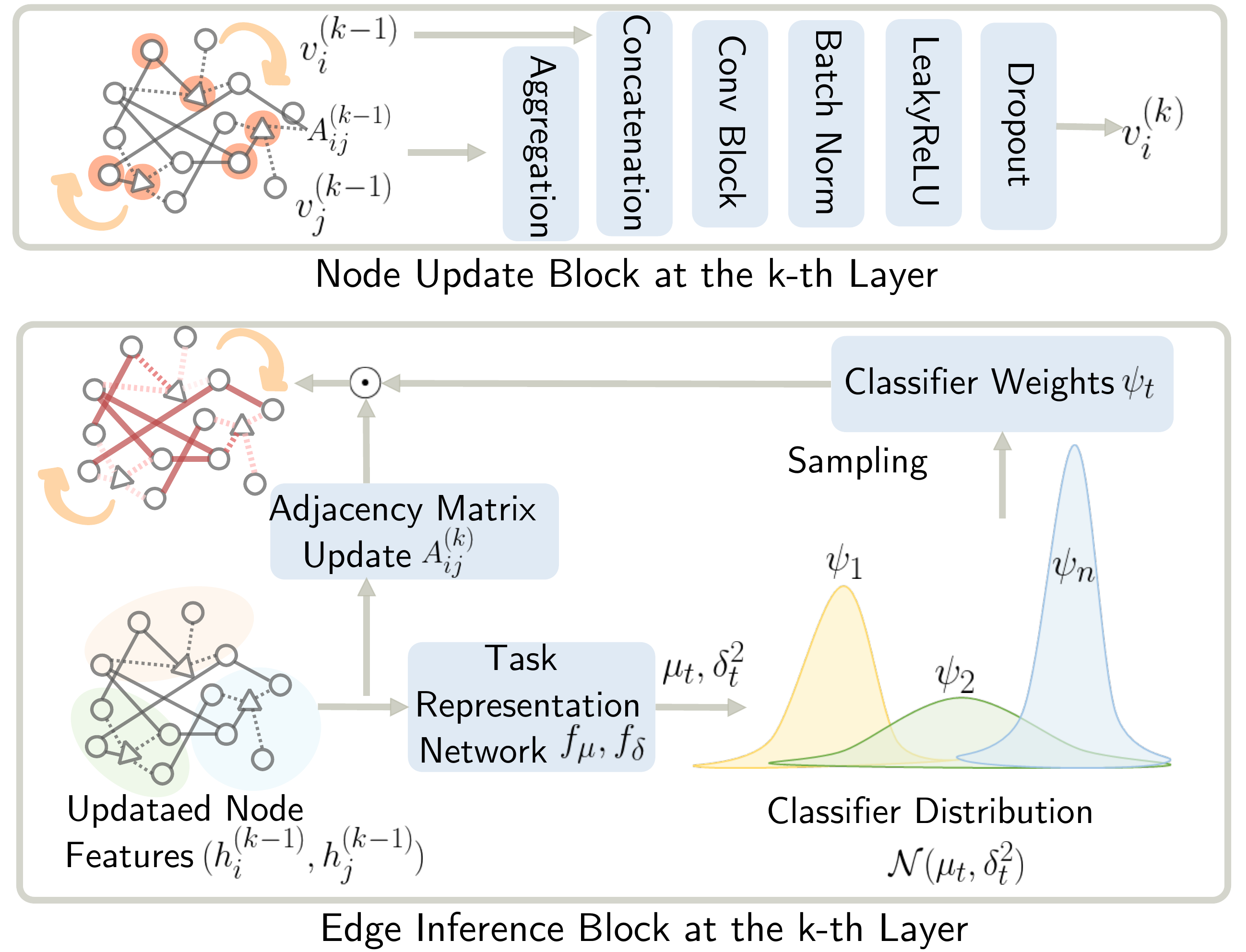}
    \caption{An illustration of the node update and edge inference block at the $k$-th layer.}
    \label{fig:node}
\end{figure}

\section{Proposed Approach}
In this section, we detail two major components of the proposed method, the continual graph neural networks and Bayes by Backprop procedure for edge inference.
\subsection{Continual Graph Neural Networks}
As shown in Figure \ref{fig:flowchart}, the support $\Tau = (\mathcal{S}, \mathcal{Q})$  (shown in circle) and query data $\mathcal{Q} = \{(\tilde{x}_i, \tilde{y}_i)\}_{i=1}^{K}$ (shown in triangle) in each episode can form an undirected acyclic graph $\mathcal{G} = (\mathcal{V}, \mathcal{E})$. Each vertex $v_i\in \mathcal{V}$ is associated with a node embedding vector $\mathcal{X}=\{x_i\}_{i=1}^{|\mathcal{S}|+|\mathcal{Q}|}$ and each edge $e_{ij}\in \mathcal{E}$ represents interaction between nodes. The adjacency matrix $\mathcal{A} = \{A_{ij}\}_{i,j=1}^{|\mathcal{V}|}$ is defied as semantic similarity between two connected nodes $v_i$ and $v_j$, which will be updated dynamically. Considering the given label of support set, the adjacency matrix can be initialized as,
\begin{equation}
    A_{ij}^{(0)} = \begin{cases}
    1  & \text{if}\,x_i, x_j\in\mathcal{S}\,\text{and}\,y_i=y_j,\\
    0  & \text{if}\,x_i, x_j\in\mathcal{S}\,\text{and}\,y_i\neq y_j,\\
    0.5 & \text{otherwise}，
    \end{cases}
\end{equation}
where support-support pairs are initialized \textit{w.r.t} their labels and query-support pairs are softly assigned with an uncertain value. Even though ambiguity is explicitly injected, we remedy the noise via probabilistic edge inference presented in the next section. For efficient implementation, we split the meta-train $\mathcal{D}_{train}$ and meta-test dataset $\mathcal{D}_{test}$ into several sequences of episodes $\{\mathcal{G}_t\}_{t=1}^T$, which is learned by node updating (see Figure \ref{fig:node}), history transition and edge inference module consecutively, where $T$ is the length of the sequence. 

\subsubsection{Node Interactions Modeling}
To make each node aggregate more information from neighbors $K$ hops away, the proposed graph model stacks $K$ aggregation blocks. Following the generic propagation rule, the node vectors in arbitrary graph $\mathcal{G}_t$ at the $k$-th layer can be updated as,
\begin{equation}\label{eq:node}
\begin{split}
    \hat{v}_i^{(k-1)}&= \sum_{j\in\mathcal{N}(i)}(v_i^{(k-1)} A^{(k-1)}_{ij}),\\
    v_i^{(k)} &= f_n([v_i^{(k-1)}; \hat{v}_i^{(k-1)}];\theta_n),
\end{split}
\end{equation}
where $\mathcal{N}(i)$ indicates the neighbor set of the node $v_i$, $[\cdot;\cdot]$ is the concatenation operation and $f_n(\cdot;\theta_n)$ is a transformation block consisting of two convolutional layers, one LeakyReLU activation and one dropout layer. The node embedding is initialized with the extracted representation from the backbone embedding model, \textit{i.e.,} $v_i^{(0)}=x_i$. Here we leave out the layer mark $^{(k)}$ in the next section for simplicity.

\subsubsection{Task History Transition}
After receiving messages from current episode $\mathcal{G}_t$, each node embedding $v^t_i$ is further transformed with a gate updater. Different from the Gated Graph Neural Network~\cite{GGNN} that utilizes the gate updater to extend the depth of graph neural networks for the same batch of data, we feed different embedding at each time step in order to capture the long-term task correlations. Specifically, the hidden state for each node is transferred as follows,
\begin{equation}\label{eq:GRU}
    \begin{split}
        z_i^{t} &= \sigma(W_z v_i^t + U_z h_i^{t-1} + b_z),\\
        r_i^t &= \sigma(W_r v_i^t + U_r h_i^{t-1} + b_r),\\
        \tilde{h}^t_i &= \text{tanh}(W_h v_i^t + U_h(r_i^t\odot h_i^{t-1}) + b_h),\\
        h_i^t &= \tilde{h}_i^t\odot z_i^t + h_i^{t-1}\odot(1-z_i^t),
    \end{split}
\end{equation}
where $h_i^t$ is the updated feature of node $v_i^t$, $\sigma$ is the sigmoid function, $W_z$, $U_z$, $W_r$, $U_r$, $W_h$, $U_h$ are learnable weights, and $b_z$, $b_r$, $b_h$ are biases of the updating function. $z_i^t$ and $r_i^t$ are update gate vector and reset gate vector, respectively. Here we let all parameters as $\theta_h$ for shorthand. The hidden state $h_i^0$ is initialized as a zero vector.

\subsubsection{Adjacency Feature Update}
After $T$-step hidden transition, we can obtain a set of final node representations at the $k$-th layer. In this way, the adjacency features $A_{ij}$ at the $k$-th layer is calculated as,
\begin{equation}\label{eq:adjacency}
\begin{split}
    \widetilde{A}_{ij}^{(k)} &= f_e(\norm{h_i^{(k-1)} - h_j^{(k-1)}}; \theta_e),\\
    A_{ij}^{(k)} &= D^{-{\frac{1}{2}}}\widetilde{A}^{(k)}_{ij}D^{-{\frac{1}{2}}} ,
\end{split}
\end{equation}
where $D$ is the degree matrix of adjacency matrix, $f_e(\cdot; \theta_e)$ is the non-linear transformation network parameterized by $\theta_e$, which includes four convolutional blocks, a batch normalization, a LeakyReLU activation and a dropout layer.

\subsection{Bayes by Backprop for Edge Inference}
Though multiple graph aggregation layers, we aim to infer a full predictive distribution over the unknown query labels relying on distributional Bayesian Decision Theory~\cite{pmaml,versa}. Notably, we amortize the posterior distribution of the classification weights $p(\psi)$ as $q_{\psi}$ to enable quick prediction at the meta-test stage and learn parameters by minimizing the average expected loss over tasks \textit{i.e.,} $\psi^{*} = \argmin_{\psi}\mathcal{L}_{B}$, where  
\begin{equation}
\begin{split}
    \mathcal{L}_{B}&=\E[\infdiv{p(\tilde{y}|\tilde{x}, \mathcal{S})}{q_{\psi}(\tilde{y}|\tilde{x}, \mathcal{S})}],\\
   \psi^{*} &= \argmax_{\psi}\E[\log\int p(\tilde{y}|\tilde{x},\psi)q_{\psi}(\psi|\tilde{x},\mathcal{S})\diff\psi].
\end{split}
\end{equation}
To ensure likelihood tractable, we use a factorized Gaussian distribution for $q_{\psi}(\psi|\tilde{x},\mathcal{S};\Theta)$ with means and variances set by the amortization network, 
\begin{equation}
    \begin{split}
         q_{\psi}(\psi|\tilde{x}, \mathcal{S};\Theta) =
         \prod_{t=1}^{|\Tau|}q_{\psi}(\psi_{t}&|\{h_i\}_{i=1}^{|\Tau|})
         = \prod_{t=1}^{|\Tau|}\mathcal{N}(\mu_{t}, \delta_{t}^2), \\
        \mu_{t} = f_{\mu}(\{h_i\}_{i=1}^{|\Tau|}; \theta_{\mu}),&~\delta_{t}^2 = f_{\delta}(\{h_i\}_{i=1}^{|\Tau|}; \theta_{\delta}).
    \end{split}
\end{equation}
With the generated posterior distribution, we can adaptively transform the predictive logits from the $k$-th layer based on the learned adjacency matrix and sampled classifier weights $\psi_t = \{W_t, b_t\}$ for each specific task, 
\begin{equation}
    \begin{split}
    p(\tilde{y}^{(k)}|\tilde{x}, h^{(k)}, \psi_t)& = M_q\odot\sigma(W_{t}A^{(k)}+b_{t})\odot M_s,\\
    W_{t},~b_{t}&\sim \mathcal{N}(\mu_{t}, \delta^2_{t}),  
    \end{split}
\end{equation}
where $\sigma$ and $\odot$ indicate the sigmoid function and Hadamard product, respectively. $M_q$ indicates the mask matrix to select query predictions, where the value is assigned $1$ when the row index belongs to query set and $0$ otherwise. Similarly, $M_s$ selects columns when the column index belongs to support set.

During the meta-training time, the proposed model is optimized by minimizing the binary cross-entropy loss of query edges and Bayes by Backprop loss, \textit{i.e.,}
\begin{equation}\label{eq:loss}
\begin{split}
    \mathcal{L}_E = -\sum_{k=1}^{K}\sum_{i=1}^{|\mathcal{Q}|} \tilde{y}_i \log (p(\tilde{y}^{(k)}_i))& + (1-\tilde{y}_i)\log(1-p(\tilde{y}^{(k)}_i)),\\
    \Theta^* = \argmin_\Theta \mathcal{L}_E& + \gamma\mathcal{L}_B, 
\end{split}
\end{equation}
where shared parameters $\Theta = \{\theta_n, \theta_h, \theta_e, \theta_\mu, \theta_\delta\}$ are jointly optimized. the loss aggregates all predictions from different layers. The overall algorithm for meta-training is shown in Algorithm \ref{alg:1}.

\begin{algorithm}[t]
	\begin{algorithmic}[1]
		\Inputs{Task distribution of $\mathcal{D}_{train}$: $p(\Tau)$;}
		\Outputs{Model Weights: $\Theta = \{\theta_n, \theta_h, \theta_e, \theta_{\mu}, \theta_{\delta}\}$;}
		\Initialize{Hyper-parameters: $M$, $K$, $\gamma$;\\
		Adjacency matrix $A$; Hidden states $h$;\\
		Visual features $\mathcal{X}$ for all tasks;\\
		Minibatch size $m$ and learning rate $\eta$;}
        \For{M iterations}
        \State Sample batch of tasks $\Tau\sim p(\Tau)$;
        \State Message passing and update node representations in Equation~\eqref{eq:node} and Equation \eqref{eq:GRU};
        \State Compute adjacency matrix in Equation \eqref{eq:adjacency}; 
        \State Generate task-specific parameter $\mu_t$, $\delta_t$, sample $\psi_t\sim\mathcal{N}(\mu_t, \delta^2_{t})$;
        \State Compute prediction $p(\tilde{y}|\tilde{x},\psi_t)$ of query samples;
        \State Update parameters $\Theta$ by descending stochastic gradients according to Equation~\eqref{eq:loss}.
        \EndFor
	\end{algorithmic}
	\caption{Meta-training of the Proposed CML-BGNN.}
	\label{alg:1}
\end{algorithm}


\section{Experiments}
\subsection{Datasets}
For fair comparisons with state-of-the-art baselines, we conduct extensive experiments on two benchmark few-shot classification datasets: 

\textbf{\textit{mini}ImageNet} is the subset of the ILSVRC-$12$ dataset, where $600$ images for each of $100$ classes are randomly chosen to be the part to the dataset. We follow the class split used by \cite{optimizationasmodel}, where $64$ classes are used for training, $16$ for validation, and $20$ for testing. All the input images have the size of $84\times 84\times 3$.

\textbf{\textit{tiered}ImageNet} is a larger subset of ILSVRC-$2012$, which contains $608$ classes in $34$ higher-level categories sampled from the high-level nodes in the ImageNet. The standard split includes $351$ classes for training, $97$ classes for validation, and $160$ classes for testing. The average number of images in each class is $1,281$.

\subsection{Baselines}
We compare our approach with the following baseline methods to justify its effectiveness:

\textit{\textbf{Optimization-based:}} Meta-learner LSTM~\cite{optimizationasmodel}, MAML~\cite{maml}, REPTILE~\cite{reptile}, Meta-SGD~\cite{meta-SGD},  SNAIL~\cite{snail}, LEO~\cite{leo}.

\textit{\textbf{Generation-based:}} PLATIPUS~\cite{pmaml}, VERSA~\cite{versa}, LwoF~\cite{dynamic}, Param\_Predict~\cite{param_prediction}, wDAE~\cite{gene_GNN}.

\textit{\textbf{Metric-based:}} Matching Net~\cite{matching}, Prototypical Net~\cite{proto}, Relation Net~\cite{relation}, TADAM~\cite{TADAM}, CTM~\cite{ctm}.

\textit{\textbf{Graph-based:}} GNN~\cite{GNN},  CovaMNet~\cite{covariance}, TPN~\cite{TPN}, EGNN~\cite{EGNN}.

\subsection{Implementation Details}\label{Implementation}
Our source code\footnote{The source code is attached in supplementary material for reference.} is implemented based on Pytorch. All experiments are conducted on a server with two GeForce GTX 1080 Ti and two GTX 2080 Ti GPUs.

\subsubsection{Module Architecture.}
Despite the generality of backbone embedding module, we adopt the same architecture used in some recent work~\cite{relation,EGNN}. Specifically, the network consists of four convolutional blocks including a 2D covolutional layer with a $3 \times 3$ kernal, a batch normalization, a $2\times2$ max-pooling and a LeakyReLU activation. Regarding to specification of recurrent units, the dimension of hidden states and all embedding size are fixed to 96. We fix the number of hidden states to 8.

\subsubsection{Parameter Settings.}
The mini-batch size for all graph-based models is 80 and 64 for 1-shot and 5-shot experiments, respectively. The proposed model was trained by Adam optimizer with an initial learning rate $\eta$ of $1\times 10^{-3}$ and weight decay of $1\times10^{-6}$. The dropout rate is set to 0.3 and the loss coefficient $\gamma$ is set to 1. We report the final results of the proposed model trained with 70K and 160K iterations on \textit{mini}ImageNet and \textit{tiered}ImageNet.

\begin{table}[t]
\caption{The 5-way 1-shot and 5-shot classification accuracies (\%) on the test split of the \textit{mini}ImageNet dataset, with $95\%$ confidence interval. $^\dag$ indicates our re-implementation. ``w/o'' indicates without.} \label{mini_result}
\resizebox{0.46\textwidth}{!}{
\begin{tabular}{r|c|cc}
\toprule
\textbf{Models} &\textbf{Backbone} &\textbf{1-shot} &\textbf{5-shot}\\
\hline
\multicolumn{4}{l}{\textbf{\textit{Optimization-based}}}\\
Meta-learner LSTM &Conv4 &$43.44\pm0.77$ &$60.60\pm0.71$\\
MAML              &Conv4 &$48.70\pm1.84$ &$63.10\pm0.92$\\
REPTILE           &Conv4 &$49.97\pm0.32$ &$65.99\pm0.58$\\
Meta-SGD          &Conv4 &$50.47\pm1.87$ &$64.03\pm0.94$\\
SNAIL             &ResNet-12&$55.71\pm0.99$ &$68.88\pm0.92$\\
LEO               &WRN-28&$61.76\pm0.08$ &$77.59\pm0.12$\\
\hline
\multicolumn{4}{l}{\textbf{\textit{Generation-based}}}\\
PLATIPUS          &Conv4 &$50.13\pm1.86$ & -\\
VERSA             &Conv4 &$53.40\pm1.82$ &$67.37\pm0.86$\\
LwoF              &Conv4 &$56.20\pm0.86$ &$72.81\pm0.62$\\
Param\_Predict    &WRN-28&$59.60\pm0.41$ &$73.74\pm0.19$\\
wDAE          &WRN-28&$61.07\pm0.15$ &$76.75\pm0.11$\\
\hline
\multicolumn{4}{l}{\textbf{\textit{Metric-based}}}\\
Matching Net      &Conv4 &$43.56\pm0.84$ &$55.31\pm0.73$\\
Prototypical Net  &Conv4 &$49.42\pm0.78$ &$68.20\pm0.66$\\
Relation Net      &Conv4 &$50.40\pm0.80$ &$65.30\pm0.70$\\
TADAM             &ResNet-12&$58.50\pm0.30$ &$76.70\pm0.30$\\
CTM               &Conv4 &$62.05\pm0.55$ &$78.63\pm0.06$\\
\hline
\multicolumn{4}{l}{\textbf{\textit{Graph-based}}}\\
GNN               &Conv4 &$50.33\pm0.36$ &$66.41\pm0.63$\\
CovaMNet          &Conv4 &$51.19\pm0.76$ &$67.65\pm0.63$\\
TPN               &Conv4 &$53.75\pm0.86$ &$69.43\pm0.67$\\
EGNN              &Conv4 &-              &$76.37\pm0.30$\\
EGNN$^{\dag}$     &Conv4 &$58.65\pm0.55$ &$75.25\pm0.49$\\
\hline
\rowcolor{Gray}\multicolumn{4}{l}{\textbf{\textit{Ours}}}\\
\rowcolor{Gray}CML-BGNN w/o C          &Conv4 &$63.74\pm0.63$ &$79.36\pm0.57$\\
\rowcolor{Gray}CML-BGNN w/o B            &Conv4 &$87.15\pm0.24$ &$91.21\pm0.19$\\
\rowcolor{Gray}\textbf{CML-BGNN}             &Conv4 &$\textbf{88.62}\pm0.43$ &$\textbf{92.69}\pm0.31$\\
\bottomrule
\end{tabular}}
\end{table}

\begin{table}[t]
\begin{center}
\caption{The 5-way 1-shot and 5-shot classification accuracies (\%) on the test split of the \textit{tiered}ImageNet dataset, with $95\%$ confidence interval.  $^\dag$ indicates the re-implementation. ``w/o'' indicates without.} \label{tier_result}
\resizebox{0.46\textwidth}{!}{
\begin{tabular}{r|c|cc}
	\toprule
	\textbf{Models} &\textbf{Backbone} &\textbf{1-shot} &\textbf{5-shot}\\
	\hline
	\multicolumn{4}{l}{\textbf{\textit{Optimization-based}}}\\
	MAML              &Conv4 &$51.67\pm1.81$ &$70.30\pm0.08$\\
	REPTILE           &Conv4 &$52.36\pm0.23$ &$71.03\pm0.22$ \\
	Meta-SGD          &Conv4 &$62.95\pm0.03$ &$79.34\pm0.06$\\
	LEO               &WRN-28&$66.33\pm0.05$ &$81.44\pm0.09$\\
	\hline
	\multicolumn{4}{l}{\textbf{\textit{Generation-based}}}\\
	LwoF              &Conv4 &$50.90\pm0.46$ &$66.69\pm0.36$\\
	wDAE          &WRN-28 &$68.18\pm0.16$ &$83.09\pm0.12$\\
	\hline
	\multicolumn{4}{l}{\textbf{\textit{Metric-based}}}\\
	Matching Net$^\dag$      &Conv4 &$54.02\pm0.00$ &$70.11\pm0.00$\\
	Prototypical Net  &Conv4 &$53.31\pm0.89$ &$72.69\pm0.74$\\
	Relation Net      &Conv4 &$54.48\pm0.93$ &$71.32\pm0.78$\\
	CTM               &Conv4 &$64.78\pm0.11$ &$81.05\pm0.13$\\
	\hline
	\multicolumn{4}{l}{\textbf{\textit{Graph-based}}}\\
	GNN              &Conv4 &$43.56\pm0.84$ &$55.31\pm0.73$\\
	TPN                     &Conv4 &$57.53\pm0.96$ &$72.85\pm0.74$\\
	EGNN-3             &Conv4 &- &$80.15\pm0.30$\\
	EGNN-1$^\dag$              &Conv4 &$61.04\pm0.45$ &$73.91\pm0.40$\\
	EGNN-2$^\dag$              &Conv4 &$64.64\pm0.50$ &$79.80\pm0.43$\\
	EGNN-3$^\dag$              &Conv4 &$65.30\pm0.53$ &$82.39\pm0.43$\\
	\hline
	\rowcolor{Gray}\multicolumn{4}{l}{\textbf{\textit{Ours}}}\\
	\rowcolor{Gray}CML-BGNN-1             &Conv4 &$85.91\pm0.51$ &$89.58\pm0.28$\\
	\rowcolor{Gray}CML-BGNN-2              &Conv4 &$88.02\pm0.50$ &$91.50\pm0.25$\\
	\rowcolor{Gray}\textbf{CML-BGNN-3}              &Conv4 &$\textbf{88.87}\pm0.51$ &$\textbf{92.77}\pm0.28$\\
	\bottomrule
\end{tabular}}
\end{center}
\end{table}

\begin{figure}[t]
    \centering
    \includegraphics[width=1\linewidth]{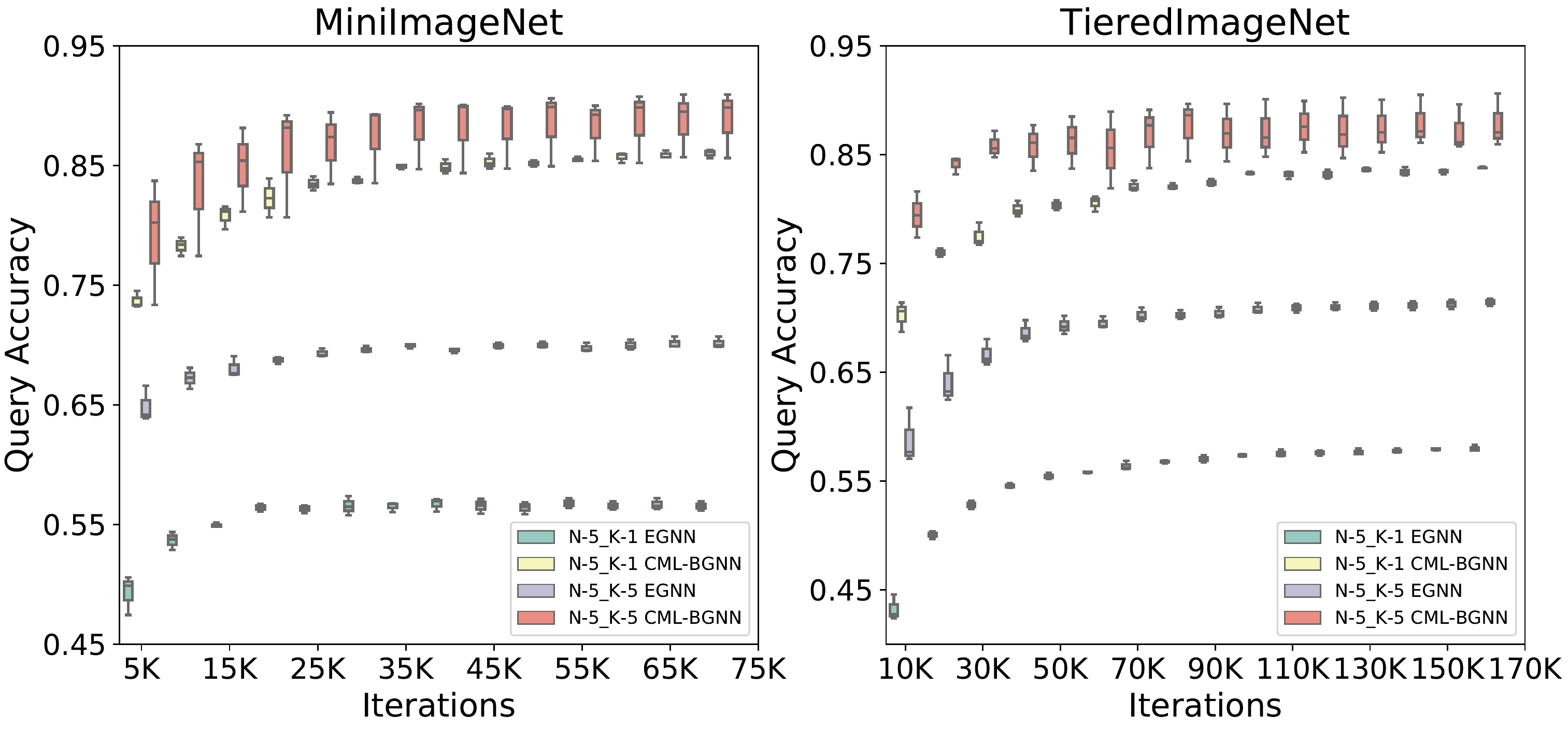}
    \caption{The 5-way 1-shot and 5-shot accuracies of query classification on the validation split of \textit{mini}ImageNet (\textit{left}) and \textit{tiered}ImageNet dataset (\textit{right}).}
    \label{fig:comp}
\end{figure}

\subsection{Comparisons with State-of-The-Art}
To verify the effectiveness of our proposed continual meta-learning model, we compare it with state-of-the-art meta-learning methods on the \textit{mini}ImageNet and \textit{tiered}ImageNet datasets. Here we report the best performance for every model in Table \ref{mini_result} and Table \ref{tier_result}, along with the specifications of the backbone embedding models for feature extraction. \textbf{Conv4} refers to a 4-layer convolutional network, \textbf{ResNet-12}~\cite{resnet} denotes 4 layer blocks of depth 3 with $3\times 3$ kernels and short connections, and \textbf{WRN-28} is a 28-layer wide residual network. Generally, a deeper embedding network will lead to a better classification performance yet with a risk of overfitting. From Table \ref{mini_result}, we can observe that our \textbf{CML-BGNN} equipped with three graph layers surpasses all compared meta-learning methods with a large margin, especially in the challenging scenario of 1-shot learning. More concretely, the proposed model with the basic conv4 embedding structure gains $43.5\%$, $45.1\%$, $42.8\%$, $51.1\%$ relative improvements over the previous best optimization-based \textbf{LEO}~\cite{leo}, generation-based \textbf{wDAE}~\cite{gene_GNN}, metric-based \textbf{CTM}~\cite{ctm} and graph-based methods \textbf{EGNN}~\cite{EGNN} in a 5-way 1-shot \textit{mini}ImageNet experiment, respectively. This is mainly owing to the learned history transition, which reinforces the memory of rare samples and correlations between classes. Furthermore, we re-implemented the most powerful graph-based baseline EGNN with mini-batch size of 80 for fairness and present a detailed comparison in boxplots. All parameters are randomly initialized in three trials with fixed seeds 111, 222, 333 for reproducibility. As depicted in Figure \ref{fig:comp}, the absolute value of validation accuracy either in 1-shot or 5-shot setting tends to go up as training iterations increase. The proposed method reaches the peaks at an early stage and achieves a much higher performance, yet showing the sensitivity to seed selection in the case of 5-way 5-shot classification. We infer this variance is mainly introduced by edge inference sampling, which can be alleviated by averaging predictions from multiple sampling. From Table \ref{tier_result}, we observe that our re-implemented EGNN obtains better performance (as indicated with $^\dag$) by enlarging the batch size from 40 to 80. This phenomenon consistently verifies that task correlations are more likely to contribute positively to few-shot learning.

\begin{figure}[t]
    \centering
    \includegraphics[width=1\linewidth,height=3.3cm]{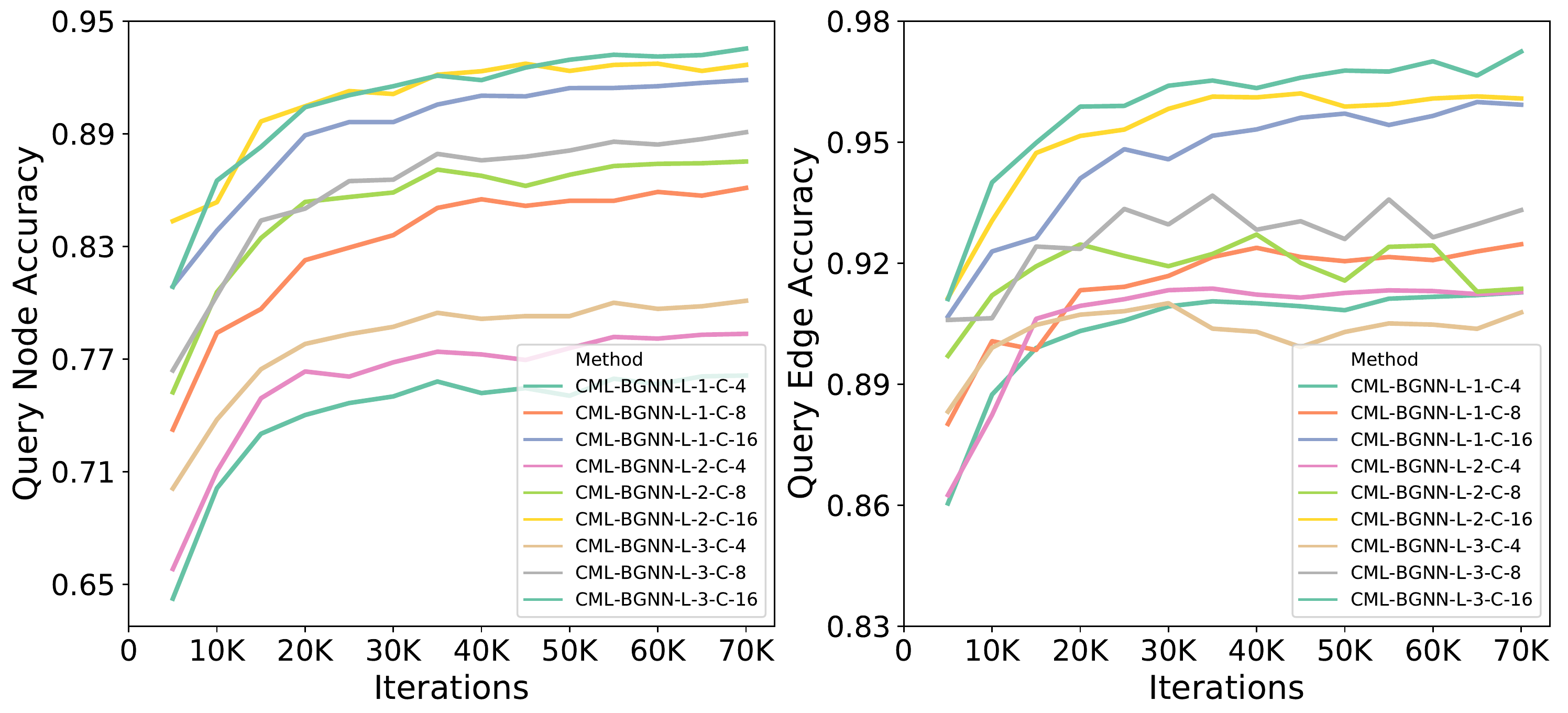}
    \caption{The 5-way 1-shot accuracies (\%) of query classification on the validation split of \textit{mini}ImageNet with different number of hidden states. Best viewed in color.}
    \label{fig:hidden}
\end{figure}

\subsection{Ablation Study}
\eat{In this section, we perform several ablative studies to study the impact of derived components, depth of graph neural networks, and number of hidden states.} 
\subsubsection{Effect of Components.}
The major ablation results regarding to CML-BGNN with different components on \textit{mini}ImageNet dataset are shown in gray blocks of Table~\ref{mini_result}. All variants are trained with three graph layers, mini-batch size of 80. Removing the history transition module, the variant \textbf{CML-BGNN w/o C} can only mine the pattern from local neighborhood without maintaining related prior information for reference, thus inevitably leading to a inferior performance, e.g., averagely decreasing 5-way 1-shot performance from $88.62\%$ to $63.74\%$. \textbf{CML-BGNN w/o B} indicates the variant of our proposed model that directly utilizes the adjacency matrix to predict query labels without inferring task-specific parameters. Accordingly, the classification accuracy suffers a slight drop on both datasets, e.g., from $92.69\%$ to $91.21\%$ in a 5-way 5-shot setting, which demonstrates the necessity of the full CML-BGNN formulations.

\begin{table}[t]
	\centering
	\caption{5-Way 5-shot and 1-shot classification accuracies (\%) on \textit{mini}ImageNet dataset with different depths of graph neural networks. $^\dag$ indicates our re-implementation.}\label{tab:depth}
	\resizebox{0.47\textwidth}{!}{
    	\begin{tabular}{l|cccc}
    		\toprule
    		 &\textbf{Methods} &Layer-1 &Layer-2 &Layer-3\\\midrule
    		\multirow{3}{*}{\textbf{1-shot}} &GNN$^\dag$ &$48.25\pm0.65$ &$49.17\pm0.35$ &$50.32\pm0.41$\\
    		&EGNN$^\dag$ &$55.13\pm0.44$ &$57.47\pm0.53$ &$58.65\pm0.55$\\
    		&\cellcolor{Gray}\textbf{CML-BGNN} &\cellcolor{Gray}$\textbf{85.73}\pm0.47$ &\cellcolor{Gray}$\textbf{87.67}\pm0.47$ &\cellcolor{Gray}$\textbf{88.62}\pm0.43$\\\hline
    		\multirow{3}{*}{\textbf{5-shot}} &GNN$^\dag$ &$65.58\pm0.34$ &$67.21\pm0.49$ &$66.99\pm0.43$\\
    		&EGNN$^\dag$ &$67.76\pm0.42$ &$74.70\pm0.46$ &$75.25\pm0.49$\\
    		&\cellcolor{Gray}\textbf{CML-BGNN} &\cellcolor{Gray}$\textbf{90.85}\pm0.27$ &\cellcolor{Gray}$\textbf{91.63}\pm0.26$ &\cellcolor{Gray}$\textbf{92.69}\pm0.31$\\
    	    \bottomrule
    	\end{tabular}}
\end{table}

\subsubsection{Effect of GNN's Depth.}
In addition to the evaluation for investigating the impact of GNN's depth, we test our model in both 5-way 1-shot and 5-shot with different depth of graph neural networks on both miniImageNet (shown in Table \ref{tab:depth}) and \textit{mini}ImageNet (shown in Table \ref{tier_result}) dataset. Generally, larger depth enables node to learn from a global perspective and thus enhances the expressive power of graph neural networks. For instance, the proposed \textbf{CML-BGNN}, \textbf{EGNN} and \textbf{GNN} equipped with a 3-layer structure respectively improve the classification accuracy by $3.4\%$, $6.4\%$ and $4.3\%$ \textit{w.r.t} 5-way 1-shot classification, compared with the one with one-layer structure.

\subsubsection{Effect of Number of Hidden States.}
In order to study the impact of number of hidden states in history transition module, we compare nine variants of our model on both datasets and show validation curves in terms of node classification accuracy and edge binary classification results in Figure \ref{fig:hidden}. The \textbf{CML-BGNN-L-\{1,2,3\}-C-16} indicates the variants that leverage unrolled gated recurrent units with 16 time steps to transfer history messages with different number of graph layers, which significantly outperform other variants with fewer hidden states \textbf{CML-BGNN-L-\{1,2,3\}-C-\{4,8\}}. This confirms that the augmented memory module effectively enhances node representation learning by bridging the prior task learning regardless of the model depth.


\begin{table}[t]
	\centering
	\caption{Semi-supervised few-shot classification accuracies (\%) on \textit{mini}ImageNet with 95\% confidence intervals.}\label{tab:semi}
	\resizebox{0.47\textwidth}{!}{
		\begin{tabular}[t]{l|ccc}
			\toprule
			\multirow{2}{*}{\textbf{Methods}} &\multicolumn{3}{c}{\textbf{5-way 5-shot}} \\ &$20\%$-labeled &$40\%$-labeled &$100\%$-labeled\\
			\midrule
			GNN-LabeledOnly &$50.33\pm0.36$ &$56.91\%\pm0.42$ &$66.41\pm0.63$ \\
			GNN-Semi &$52.45\pm0.88$ &$58.76\%\pm0.86$ &$66.41\pm0.63$ \\
			EGNN-LabeledOnly$^{\dag}$ &$58.65\pm0.55$ &$56.91\%\pm0.00$ &$75.25\pm0.49$ \\
			EGNN-Semi$^{\dag}$ &$63.62\pm0.00$ &$64.32\%\pm0.00$ &$75.25\pm0.49$\\
			\hline
			\rowcolor{Gray}\textbf{CML-BGNN-LabeledOnly} &$84.37\pm0.54$ &$88.62\%\pm0.29$ &$\textbf{92.69}\pm0.31$\\
			\rowcolor{Gray}\textbf{CML-BGNN-Semi} &$\textbf{88.95}\pm0.32$ &$\textbf{89.70}\%\pm0.32$ &$\textbf{92.69}\pm0.31$\\
			\bottomrule
	\end{tabular}}
\end{table}%
\subsection{Robustness Evaluation by Semi-supervised Learning}
To quantitatively analyze the model capacity of handling uncertainty, we conduct 5-way 5-shot semi-supervised experiments on \textit{mini}ImageNet dataset and showcase major results in Table \ref{tab:semi}. In this semi-supervised regime, support data is partially labeled while balanced across all classes, which poses a greater challenge of modeling uncertain relationships between labeled and unlabeled samples. In particular, the \textbf{$20\%$-labeled} column indicates that each episode contains 4 labeled support instances and 1 unlabeled instance. Here we use \textbf{LabeledOnly} to denote the strategy with only labeled support samples, and \textbf{Semi} presents training with both labeled and unlabeled data. By comparing the results with all graph-based counterparts, the proposed method greatly outperforms with a large margin ($88.95\%$ vs $63.62\%$ and $52.45\%$, when 20\% are labeled). The superior performance results from our uncertainty modeling, which effectively adapts the noise and misguidance from adjacency initialization with task-specific parameters.

\section{Conclusion}
In this work, we propose a continual meta-learning model with Bayesian graph neural networks for few-shot classification problem. Towards preserving more history messages associated with related tasks, the proposed CML-BGNN mines the prior knowledge patterns by updating a memory-augmented graph neural network and handles topological uncertainty with Bayes by Backprop. Distinguishing our work from conventional graph-based meta-learning methods, it naturally alleviates the catastrophic forgetting and insufficient robustness issues and thus encourages an efficient adaptation and generalization to novel tasks.

\bibliography{aaai.bib}
\bibliographystyle{aaai}

\end{document}